\documentclass{article}

\usepackage{microtype}
\usepackage{graphicx}
\usepackage{multicol}
\usepackage{dblfloatfix}

\usepackage{subfigure}
\usepackage{booktabs} % for professional tables

\usepackage{hyperref}

\usepackage[accepted]{icml2024}

\usepackage{amsmath}
\usepackage{amssymb}
\usepackage{mathtools}
\usepackage{amsthm}
\usepackage{bm}

\usepackage[capitalize,noabbrev]{cleveref}

\theoremstyle{plain}

\theoremstyle{definition}

\theoremstyle{remark}

\usepackage[textsize=tiny]{todonotes}

\icmltitlerunning{Towards General Neural Surrogate Solvers with Specialized Neural Accelerators}

\begin{document}

\twocolumn[
\icmltitle{Towards General Neural Surrogate Solvers with Specialized Neural Accelerators}

% It is OKAY to include author information, even for blind
% submissions: the style file will automatically remove it for you
% unless you've provided the [accepted] option to the icml2024
% package.

% List of affiliations: The first argument should be a (short)
% identifier you will use later to specify author affiliations
% Academic affiliations should list Department, University, City, Region, Country
% Industry affiliations should list Company, City, Region, Country

% You can specify symbols, otherwise they are numbered in order.
% Ideally, you should not use this facility. Affiliations will be numbered
% in order of appearance and this is the preferred way.
\icmlsetsymbol{equal}{*}

\begin{icmlauthorlist}
\icmlauthor{Chenkai Mao}{stanford}
\icmlauthor{Robert Lupoiu}{stanford}
\icmlauthor{Tianxiang Dai}{stanford}
\icmlauthor{Mingkun Chen}{stanford}
\icmlauthor{Jonathan A. Fan}{stanford}
\end{icmlauthorlist}

\icmlaffiliation{stanford}{Department of Electrical Engineering, Stanford, Palo Alto, USA}

\icmlcorrespondingauthor{Jonathan Fan}{jonfan@Stanford.edu}

% You may provide any keywords that you
% find helpful for describing your paper; these are used to populate
% the "keywords" metadata in the PDF but will not be shown in the document
\icmlkeywords{Surrogate Solver, PDE, neural operator, domain decomposition}

\vskip 0.3in
]

% this must go after the closing bracket ] following \twocolumn[ ...

% This command actually creates the footnote in the first column
% listing the affiliations and the copyright notice.
% The command takes one argument, which is text to display at the start of the footnote.
% The \icmlEqualContribution command is standard text for equal contribution.
% Remove it (just {}) if you do not need this facility.

\printAffiliationsAndNotice{}  % leave blank if no need to mention equal contribution
% \printAffiliationsAndNotice{\icmlEqualContribution} % otherwise use the standard text.

\begin{abstract}
Surrogate neural network-based partial differential equation (PDE) solvers have the potential to solve PDEs in an accelerated manner, but they are largely limited to systems featuring fixed domain sizes, geometric layouts, and boundary conditions. We propose Specialized Neural Accelerator-Powered Domain Decomposition Methods (SNAP-DDM), a DDM-based approach to PDE solving in which subdomain problems containing arbitrary boundary conditions and geometric parameters are accurately solved using an ensemble of specialized neural operators.  We tailor SNAP-DDM to 2D electromagnetics and fluidic flow problems and show how innovations in network architecture and loss function engineering can produce specialized surrogate subdomain solvers with near unity accuracy.  We utilize these solvers with standard DDM algorithms to accurately solve freeform electromagnetics and fluids problems featuring a wide range of domain sizes. Code for this project could be found at: \url{https://github.com/ChenkaiMao97/SNAP-DDM}
\end{abstract}

\section{Introduction}
Large scale physics simulations are critical computational tools in every modern science and engineering field, and they involve the solving of parametric partial differential equations (PDEs) with different physical parameters, boundary conditions, and sources.  Their ability to accurately capture underlying physical processes makes them particularly well suited in modeling and optimization tasks.  Conventionally, PDE problems are set up using discretization methods such as the finite element or finite difference formalisms, which frame the PDE systems as large sparse matrices that are solved by matrix inversion.  Problems are set up from scratch every time and computational scaling with domain size is fundamentally tied to the scaling of matrix inversion algorithms.
%, and they are utilized in applications as diverse as aircraft modeling, climate prediction, and nanomaterials design
%, mandating considerable time and computational resource requirements for large scale simulations.

%utilization of qualitatively different mathematical approaches to solving PDEs, which have the
Neural network-based approaches to solving PDE problems have emerged and have garnered great interest due to their tantalizing potential to exceed the capabilities of conventional algorithms.  One of the earliest and most prominent concepts is the Physics Informed Neural Network (PINN), which produces an ansatz for a given PDE problem \cite{raissi2019physics, karniadakis2021physics, cai2021physics}. PINNs have been shown to be able to solve wave propagation problems with fixed domain size and domain geometry, but their accuracy is sub-optimal \cite{moseley2020solving, rasht2022physics} in systems featuring high spatial frequency phenomena \cite{wang2022and, farhani2022momentum}. In addition, they require retraining every time the PDE problem is modified, making them unsuitable for solving generalized parametric PDE problems.
%This mesh free method uses the governing PDE equations and boundary conditions to construct the loss function for deep network training, and it does not require training data. 
%The main limiting factor for PINNs is that the network needs to be re-trained for each new problem, and it usually extrapolates poorly to unseen regions. These disadvantages make it hard to apply PINNs for large-scale general problems.

% One of the challenge in training PINNs is to balance different physical loss terms. Wang et.al. presents a learning rate annealing algorithm that utilizes gradient statistics during model training
% to help balance different terms\cite{wang2021understanding}.

%demonstrated their ability to solve PDEs and related inverse design problems with potential to improve efficiency \cite{raissi2018deep, kochkov2021machine, chen2022wavey, sanchez2020learning}. 
Neural Operators, which are the focus of this study, have also been recently proposed as deep network surrogate PDE solvers.  Unlike PINNs, Neural Operators learn a family of PDEs by directly learning the mapping of an input, such as PDE coefficients, to corresponding output solutions using simulated training data. PDE solutions are evaluated through model inference, as opposed to model training, which enables exceptionally high speed PDE problem solving.  Initial work on Neural Operator models can be traced to PDE-Net \cite{long2018pde, long2019pde}, and additional improvements in network architecture have been proposed with DeepONet \cite{lu2019deeponet} and Fourier Neural Operators (FNO) \cite{li2020fourier}.  While much progress has been made, Neural Operators cannot yet directly scale to large arbitrary domain sizes, and they cannot accurately handle  arbitrary boundary conditions.  These challenges arise due to multiple reasons: 1) the dimensionality of PDE problems grows exponentially with problem scale and can outpace the expressiveness of deep neural networks; 2) it remains difficult to scale neural networks to large numbers of parameters; and 3) the large scale generation of training data for the training of large scale models is  resource consuming.
%most Neural Operator models remain limited to small-scale toy model problems with fixed boundary conditions. 
%Initial work on Neural Operator models can be traced to PDE-Net, which introduced constrained filters in convolutions to learn the operator and predict dynamic responses\cite{long2018pde, long2019pde} within convection-diffusion problems.  Additional improvements in network architecture were proposed with DeepONet\cite{lu2019deeponet}, which used branch and trunk sub-networks to learn operators for different input and output meshes, and the Fourier Neural Operator (FNO)\cite{li2020fourier}, which parameterizes the integral kernel directly in Fourier space using Fast Fourier Transforms (FFTs) and can achieve zero-shot super-resolution.  

%While much progress has been made, scaling Neural Operators to large practical problems of arbitrary size with heterogeneous parameters and arbitrary boundary conditions remains challenging due to multiple reasons: 1) The dimensionality of PDE problems grows exponentially with problem scale and can outpace the expressiveness of deep neural networks; 2) , limiting their ability to accurate model the solutions to large scale problems.  Second, outside of transformer-based architectures recently used in large language models, it remains difficult to scale neural networks to large numbers of parameters. Third, the large scale generation of training data for the training of large scale models is resource consuming and undesirable.

\begin{figure*}[!b]
  \centering
  \includegraphics[scale=0.5]{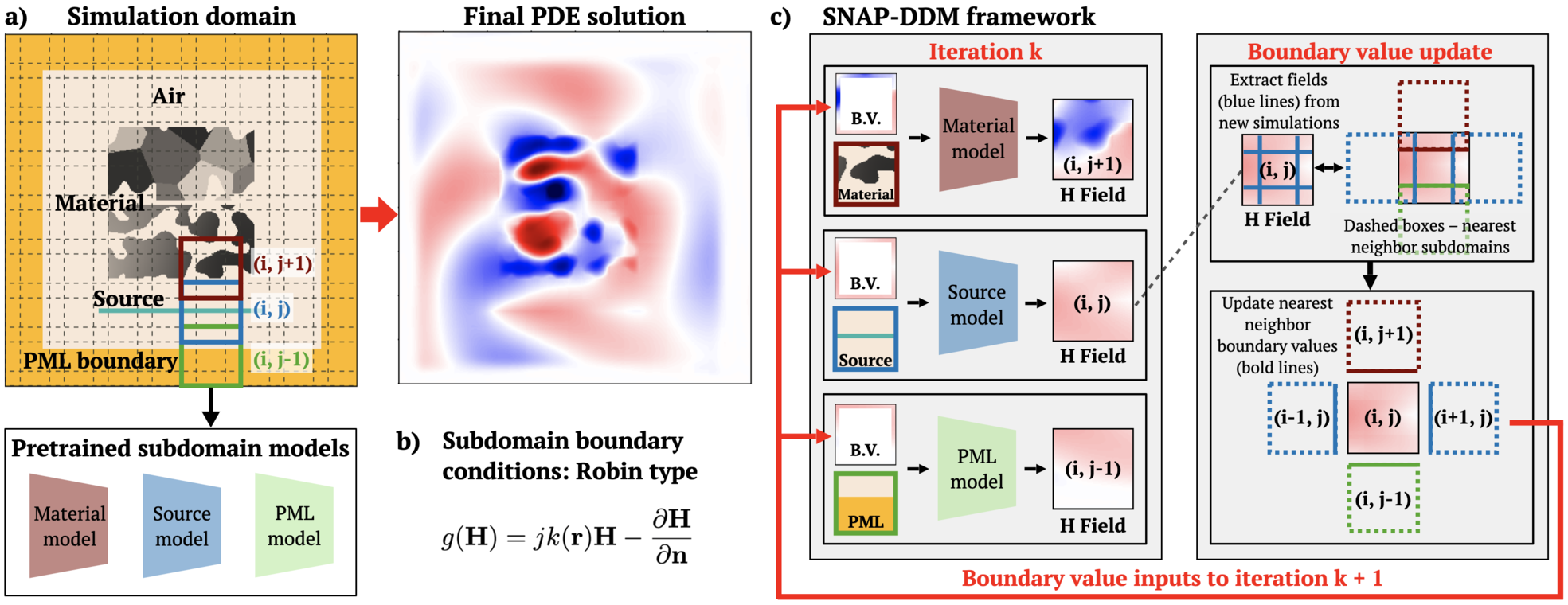}
  \caption{SNAP-DDM framework, with Electromagnetics as a demonstration. a) Global simulation domain and corresponding H-field solution for a 2D electromagnetics problem featuring arbitrary sources, global boundary conditions, and freeform grayscale dielectric structures.  The global domain is subdivided into overlapping subdomains parameterized by position $(i,j)$. Three types of specialized Neural Operator models are trained to solve for three types of subdomain problems.  b) Expression for the Robin type boundary condition used in the specialized Neural Operator subdomain models. $k(\mathbf{r}) = 2\pi \varepsilon(\mathbf{r})/\lambda$ is the wave vector in a medium with dielectric constant $\varepsilon$ and $\mathbf{n}$ is the outward normal direction.  c) Flow chart of the iterative overlapping Schwarz method.  In iteration $k$, electromagnetic fields in each subdomain are solved using the specialized Neural Operators, and the resulting fields are used to update the subdomain boundary value inputs for iteration $k+1$.  The ``Boundary value update" box shows how solved fields in the $(i,j)$ subdomain are used to update the boundary value fields in nearest neighbor subdomains for subsequent iterations.  
  \textbf{B.V.}: boundary value. \textbf{PML}: perfectly matched layers. }\label{fig-1}
\end{figure*}

In this work, we propose Specialized Neural Accelerator-Powered Domain Decomposition Methods (SNAP-DDM), which is a qualitatively new way to implement Neural Operators for  solving large scale PDE problems with arbitrary domain sizes and boundary conditions.   Our method circumvents the issues posed above by subdividing global boundary value problems into smaller boundary value subdomain problems that can be tractably solved with Neural Operators, and then to stitch together subdomain solutions in an iterative, self-consistent method using Domain Decomposition Methods (DDMs).  DDMs are the basis for solving large PDE problems with parallel computing \cite{smith1997domain, dolean2015introduction}, and they can be implemented using various algorithms including the Schwarz, finite-element tearing and interconnecting \cite{wolfe2000parallel}, optimized Schwarz  \cite{gander2002optimized}, two-level \cite{farhat2000two}, and sweeping preconditioner \cite{poulson2013parallel} methods. We leave more discussions and references of DDM in Appendix \ref{appendix:DDM-reference} for interested readers. While DDM methods have been previously explored in the context of PINNs \cite{jagtap2020conservative,jagtap2021extended}, the accurate solving of arbitrary PDE problems using the combination of Neural Operators and DDM has not been previously reported. 
%To illustrate the concept, we implement SNAP-DDM to solve two-dimensional electromagnetics and fluid mechanics problems featuring varying domain dimensions and content. 

A principal challenge in adapting Neural Operators to DDM is that the subdomain solvers require exceptional accuracy and generalizability to enable accurate DDM convergence \cite{corigliano2015model}.   To address this challenge, we train specialized Neural Operators that each solve particular classes of subdomain problems, such as those containing only sources or structural geometric parameters as model inputs.  We also propose the Self-Modulating Fourier Neural Operator (SM-FNO) architecture, an augmented FNO architecture with modulation connections that is capable of learning complex PDE boundary value operators with over 99\% accuracy.  We integrate these Neural Operators directly into a Schwarz DDM iterative framework, where field solutions within each subdomain are iteratively solved until the fields in and between every subdomain are self-consistent, at which point the global field solutions are converged.  
%With these concepts, our trained subdomain solvers are over 99\% accurate.  
% for solving PDE problems with arbitrary domain size and high accuracy.  In this DDM method,

%We summarize our contributions as follows:
%\begin{itemize}
%\item \textbf{We propose a general framework} for scaling neural network surrogate solvers to large domains by training an ensemble of subdomain boundary value PDE solvers and utilizing them directly within domain decomposition algorithms.
%\item \textbf{We propose the Self-Modulating Fourier Neural Operator (SM-FNO)}, an augmented FNO architecture with modulation connections that is capable of leaning complex PDE operators and enabling accurate, specialized semi-general subdomain neural surrogate solvers.
%\item \textbf{We release \emph{Maxwell-BC-2D}}, a subdomain boundary value problem dataset for 2D Maxwell's Equation, which contains 1M data with highly heterogeneous dielectric distributions and semi-general boundary conditions. Sources and PMLs are also included for full-problem learning. To our knowledge, this is the first general Maxwell's Equation dataset with semi-general boundary conditions and source and PMLs.
%\item \textbf{We release \emph{Maxwell-DDM}}, a large scale full-wave simulation dataset for both 2D and 3D Maxwell's Equation with highly heterogeneous dielectric distributions, sources and PMLs. This serves as a testbed for trained SNAP-DDM algorithms.
%\end{itemize}

\section{Methods}
%\subsection{Electromagnetics}

For this study, we will initially focus on classical electromagnetics (EM) as a model system for detailed analysis, followed by demonstrations of SNAP-DDM to fluid mechanics problems.  Classical EM PDEs are governed by Maxwell's equations.  The frequency domain magnetic field wave equation is:
\begin{equation}\label{eq-1}
\nabla \times (\frac{1}{\varepsilon(\mathbf{r})} \nabla \times \mathbf{H}(\mathbf{r})) - \mu_0 \omega^2 \mathbf{H}(\mathbf{r}) = i\omega \mathbf{J}(\mathbf{r})
\end{equation}
$\omega$ is angular frequency, $\varepsilon(\mathbf{r})$ is a heterogeneous dielectric material distribution that is a function of spatial position $\mathbf{r}$, $\mathbf{J}(\mathbf{r})$ is the current source distribution, and $\mathbf{H}(\mathbf{r})$ is the magnetic field distribution to be solved.  

\begin{figure*}[!b]
  \centering
  \includegraphics[scale=0.28]{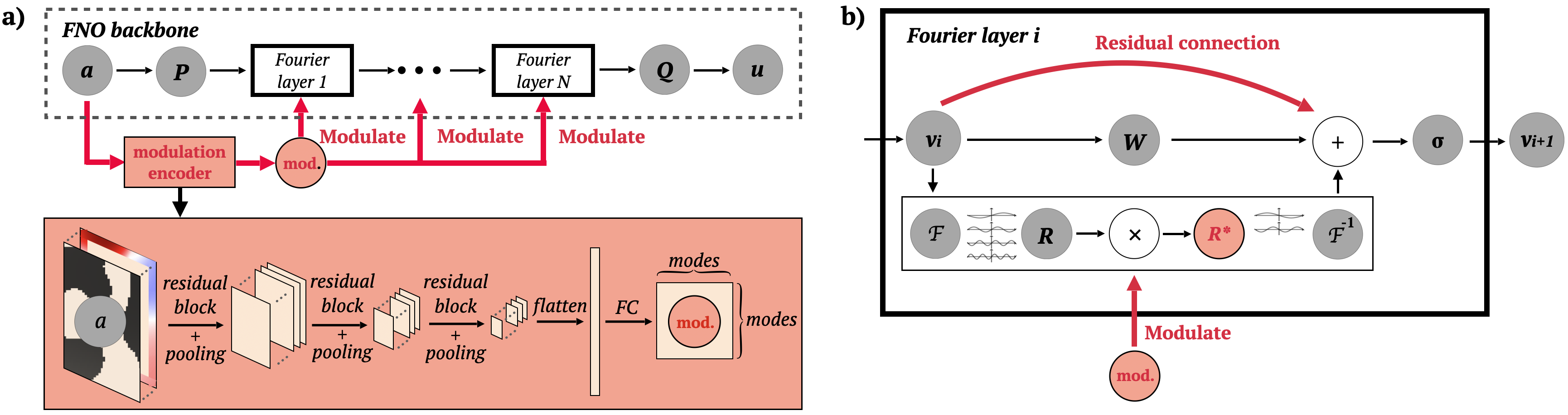}
  \caption{Self-Modulating Fourier Neural Operator architecture for DDM subdomain solvers.  Modifications to the standard Fourier Neural Operator (FNO) are highlighted in red and include: 1) the addition of self-modulation connections that encode the input into a tensor, which is then multiplied with the linear transformation matrix \textbf{R*} in each Fourier layer; and 2) the addition of residual connections inside each Fourier layer.  \textbf{a}: network input comprising a stack of images specifying the specialized subdomain layout and Robin boundary values.  \textbf{u}: network output comprising a stack of images specifying the real and imaginary H-field maps.  \textbf{P}: fully connected layer that increases the number of channels. \textbf{Q}: fully connected layer that decreases the number of channels. $\mathcal{F}$: Fourier transform. $\sigma$: leaky-ReLU activation function \cite{xu2015empirical}. \textbf{W}: channel mixer via 1-by-1 kernel convolution. \textbf{R}: original linear transform on the lower Fourier modes. \textbf{R*}: modulated linear transform through an element-wise multiplication with the modulation tensor.}\label{fig-2}
\end{figure*}

\subsection{SNAP-DDM}
%Instead of learning an end-to-end model that approximates the PDE operator, which easily becomes intractable due to the exponentially growing complexity, we adopt a domain decomposition approach to solve the whole problem iteratively. Using overlapping Schwarz Method, we subdivide the domain into overlapping subdomains with fixed size, with each subdomain to be solved by a certain specialized neural operator. 

To solve an arbitrarily sized PDE system, we consider a domain decomposition approach where the global domain is subdivided into overlapping subdomains with fixed $64 \times 64$ grids (Figure \ref{fig-1}a).  Each subdomain for a given DDM iteration poses a boundary value problem that are solved using a specialized pretrained subdomain model.  In this study, we utilize the overlapping Schwarz DDM formalism using Robin boundary conditions for each subdomain solver (Figure \ref{fig-1}b)\cite{gander2001optimized, dolean2009optimized}.

The DDM iterative algorithm is illustrated in Figure \ref{fig-1}c and the procedure is summarized as follows:
\begin{enumerate}
    \item The field is initialized to be zero everywhere.
    \item For the $k^{th}$ iteration, a boundary value problem is solved using a specialized subdomain model at each subdomain. For each subdomain, the inputs are the subdomain Robin boundary values and a specialized image of the domain (i.e., image of the material, source, or boundary layers) and the output is the field map. 
    \item Fields outputted from the models are used to update the subdomain Robin boundaries in all subdomains, which are used as model inputs for the $(k+1)^{th}$ iteration.
    % \item The newly solved magnetic fields evaluated in the $k^{th}$ iteration are used as new boundary conditions for the neighboring subdomains
    \item The algorithm terminates when a predetermined number of iterations is executed or when the physics residue falls below a predetermined threshold.
\end{enumerate}  
%in which the appropriate specialized neural operators are applied
  
It is essential that the trained subdomain PDE surrogate solvers have near unity accuracy to ensure that the DDM algorithm accurately converges.  Therefore, we train specialized neural operators that each solve specific classes of PDE problems.  For 2D EM problems, we consider three types of Neural Operators that each specialize in solving: 1) subdomains containing only PMLs in air; 2) subdomains containing only sources in air; and 3) subdomains containing only heterogeneous grayscale material and air structures. In this way, we partition complex PDE systems into regions with similar physical characteristics, which reduces the dimensionality of the learning problem and enables specialized physics to be more accurately captured in each network. Additional specialized neural operators can be considered with increasing problem domain complexity without loss of generality.  

\begin{table*}[!hb]
  \caption{Electromagnetics: Subdomain model benchmark on 10k test data}
  \label{sub-bench-table}
  \vskip 0.05in
  \begin{center}
  \begin{small}
 .\begin{sc}
  \begin{tabular}{llllll}
  \toprule
    Model (trained on 100k data)  & $L_{data}$ (\%) & $L_{pde}$ (a.u.)  & $L_{bc}$ (a.u.)      & Param (M) & FLOP (G)     \\ 
    \midrule
    FNO-v1         &  9.04  & 2.21  & 0.309  & 73.8  & 0.79 \\
    F-FNO-v1       &  8.32 & 1.69  & 0.163  & 4.9   & 1.45 \\
    UNet-v1        &  5.40  & 0.73  & 0.099  & 5.2  & 1.53 \\
    Swin T-v1      &  5.15  & 2.15  & 0.148  & \textbf{1.9}   & 9.60\\
    SM-FNO-v1-data-only(ours)& 3.95  &  7.08 & 0.162  & 4.7   & \textbf{0.66}  \\
    SM-FNO-v1(ours)&  \textbf{3.85}  & \textbf{0.50}  & \textbf{0.067}  & 4.7   & \textbf{0.66}  \\
    \midrule
    Model (trained on 1M data)  & $L_{data}$ (\%) & $L_{pde}$ (a.u.)  & $L_{bc}$ (a.u.)      & Param (M) & FLOP (G)     \\ 
    \midrule
    FNO-v2         &  5.34  & 1.43  & 0.124  & 131.2 & 1.86 \\
    F-FNO-v2       &  3.52  & 0.84  & 0.078  & 13.3  & 2.59 \\
    UNet-v2        &  2.93  & 0.44  & 0.080  & 11.1  & 3.28 \\
    SM-FNO-v2-data-only(ours)&  1.36 & 2.76  & 0.073 &  \textbf{10.2} & \textbf{1.43}  \\
    SM-FNO-v2(ours) & \textbf{1.01} & \textbf{0.30} & \textbf{0.030} & \textbf{10.2} & \textbf{1.43} \\
    \bottomrule
  \end{tabular}
  \end{sc}
  \end{small}
  \end{center}
  \vskip -0.1in
\end{table*}

\subsection{Self-Modulating Fourier Neural
Operator}
For the subdomain models, we modify the original FNO architecture \cite{li2020fourier} and introduce the Self-Modulating Fourier Neural Operator (SM-FNO) subdomain surrogate solver (Figure \ref{fig-2}).  We specifically incorporate two key features, the first of which we term a modulation encoder. Mechanistically, we utilize multiple residual blocks and fully connected layers to compress the input data into a latent modulation tensor, which then modulates each linear transformation \textbf{R} in the neural operator through element-wise multiplication (Figure \ref{fig-2}). This concept builds on our observation that in the original FNO architecture, the linear transform weight \textbf{R} in each Fourier layer is fixed and are independent of the network input parameters, limiting the ability of the neural operator to accurately process the highly heterogeneous input data featured in our problem.  This modification is inspired by the efficacy of self-attention in transformer architectures \cite{vaswani2017attention} and multiplicative interactions between inputs in PINNs \cite{wang2021understanding}. Other works have also attempted to modify the integral kernel in the FNO layers, with smoothened masking functions that mostly applies to binary input shapes\cite{liu2024domain}. With our grayscale input that represents material dielectric data, our learnable modulation offers more flexibility. We show in Section \ref{ablation} that the modulation method we introduced is crucial to enhance the expressivity of the model.

The second feature we propose is the explicit addition of a residual connection within each Fourier layer. The residual connection concept dates back to the ResNet architecture \cite{he2016deep}, where such connections were shown to mitigate the vanishing gradient problem and enable the training of deeper models. From our experiments, we have discovered that explicit residual connections are necessary for training deeper FNOs, especially when inputs are augmented with auxiliary data like boundary values. We note that the residual connection is equivalent to initializing the $1 \times 1$ convolutional layer \textbf{W} using identity plus Kaiming or Xavier initialization \cite{he2015delving}, but we keep the residual connection in Figure  \ref{fig-2} for clarity and ease of implementation. We also note that related concepts involving the addition of residual connections to the FNO architecture have been explored elsewhere \cite{tran2021factorized}.
%Although in the original architecture, the linear block $W$ can also serve as residual connections if the kernel weights are close to one, but 

\subsection{Hybrid data-physics loss function}
To train the subdomain solvers, we apply a training scheme that utilizes a hybrid data-physics loss function:
\begin{equation}\label{totloss}
    L = L_{data} + \alpha (L_{pde}+c L_{bc})
\end{equation}
% \begin{equation}\label{dataloss}
%     L_{data} = \frac{1}{N}\sum_{n=1}^{N}  \Big|\Big|\mathbf{H}^{(n)}-\hat{\mathbf{H}}^{(n)}\Big|\Big|_1 
% \end{equation}
% \begin{equation}\label{PDEloss}
%     L_{pde} = \frac{1}{N}\sum_{n=1}^{N}\Big|\Big|\nabla \times (\frac{1}{\varepsilon(\mathbf{r})} \nabla \times \mathbf{H}^{(n)}) - \mu_0 \omega^2 \mathbf{H}^{(n)}\Big|\Big|_1  
% \end{equation}
% \begin{equation}\label{PDEbc}
%     L_{bc} = \frac{1}{N}\sum_{n=1}^{N}\Big|\Big|g - \big(jk(\mathbf{r})\mathbf{H}^{(n)} -\frac{\partial \mathbf{H}^{(n)}}{\partial n}\big)\Big|\Big|_1 
% \end{equation}
Detailed expressions of the data and physics loss terms are in Appendix \ref{appendix:loss-function}. $c$ is a constant weight set to 1 for simplicity (we found the model performance is insensitive to its value between $0.1$ and $10$) and  $\alpha$ is a dynamic weighting term that balances data loss and physics loss from history loss statistics (details in Appendix \ref{appendix:physics-training}). Regularization of network training with physical loss serves to explicitly enforce physical relationships between nearest neighbor fields, which enhances the accuracy of magnetic field spatial derivatives calculated using the finite differences method \cite{chen2022wavey}.  Such accuracy is critical for evaluating electric fields and Robin boundary conditions from inferenced magnetic fields.

\section{Experiments} \label{Experiments}
%(we didn't do any parallelazation aside from numpy's BLAS library multi-thread acceleration)
\subsection{Data generation}
We used an established finite-difference frequency domain (FDFD) solver \cite{hughes2019forward} to generate 1000 full-wave simulations, each containing heterogeneous dielectric distributions (refractive index from 1 to 4) within a $960 \times 960$ grids simulation domain.  The structural layouts are specified by a pipeline inspired from image processing, with details provided in Appendix \ref{appendix:data-generation}. Random magnetic current sources surrounding the devices are specified as a superposition of sinusoidal functions with random amplitudes and phases. 40-grid-thick uniaxial PML layers are place on the 4 sides \cite{gedney1996anisotropic}. The resulting simulated fields are then cropped into $64 \times 64$ grids sections to produce the subdomain training dataset. Using this approach, we generate a total of 1.2M subdomain training data samples (100k for the PML solver, 100k for the source solver, and 1M for the grayscale material solver). 
%Equation (\ref{totloss}) write out function 
%, and $g$ is the ground truth transmission Robin boundary condition value

\begin{figure*}[!ht]
      \centering
      \makebox[\textwidth][c]{\includegraphics[scale=0.53]{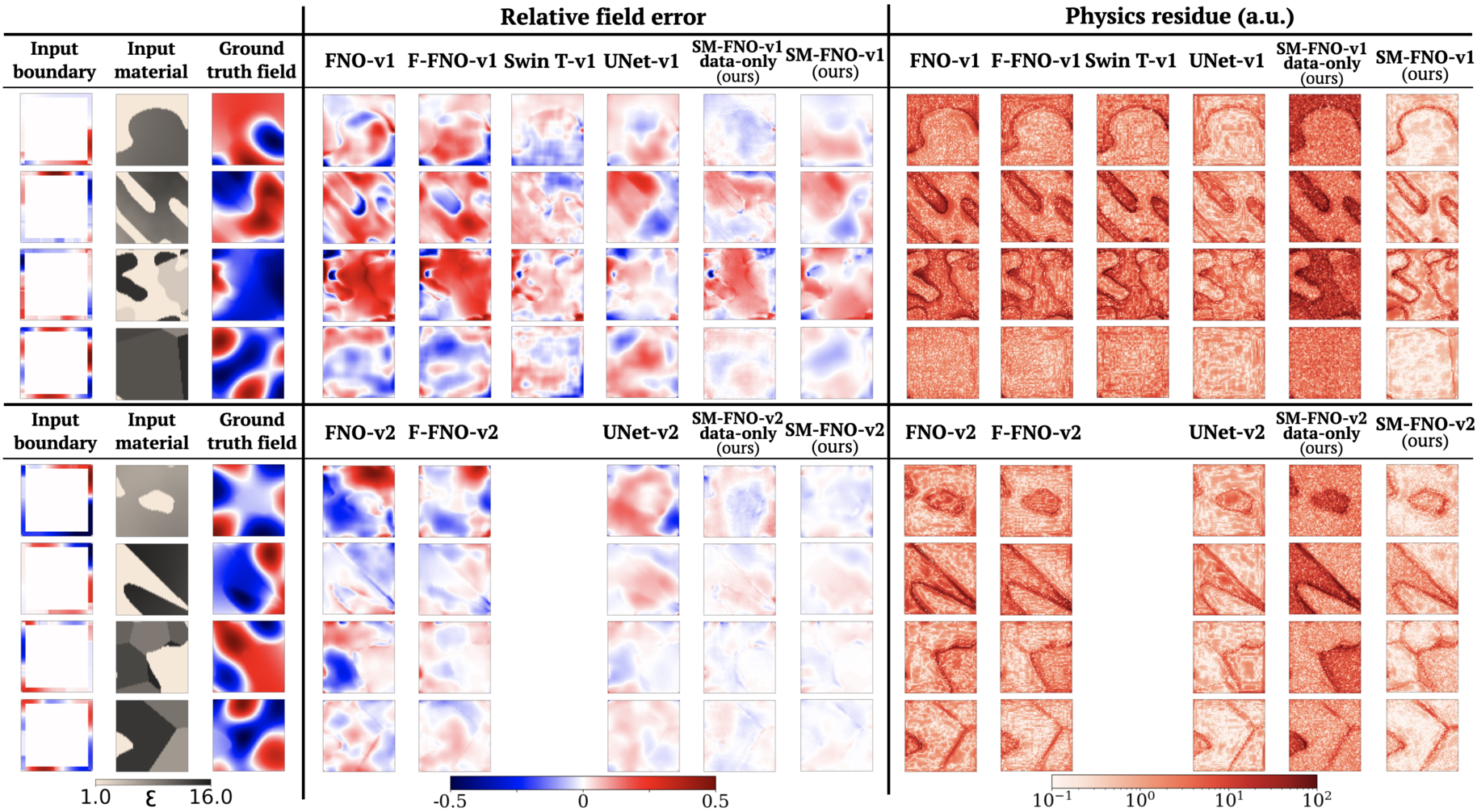}}%
      \caption{Benchmarking of material boundary value subdomain solvers on unseen test data. The model inputs are a grayscale material dielectric distribution image ($\varepsilon=1$ to $\varepsilon=16$) and Robin boundary conditions, and the outputs are images of the real and imaginary H-fields. The real parts of outputted H-fields are shown. The L1 data loss is normalized to the mean absolute ground truth field value and the physics residue map is the summed expression in Equation \ref{PDEloss}.}\label{fig-3}
\end{figure*}

\subsection{Subdomain network benchmark}
We benchmark our trained SM-FNO subdomain solver with the UNet \cite{ronneberger2015u}(details in Appendix \ref{appendix:baseline-unet}), Swin Transformer \cite{liu2021swin} (details in Appendix \ref{appendix:baseline-swint}), the classical FNO, and the recently improved version of FNO termed F-FNO \cite{tran2021factorized}. We also train our SM-FNO solver without physics loss. The networks are trained with both 100K and 1M total data to show their dependency on data scaling, except for the Swin Transformer model, which is only benchmarked on 100k training data (training on 1M data would take 2 months). The 100k and 1M data are split into 90\% training data and 10\% test data. All models use a batchsize of 64 and are trained for 100 epochs for 100k training data or 50 epochs for 1M training data. The Adam optimizer with an individually fine-tuned learning rate is used with an exponential decay learning rate scheduler that decreases the learning rate by 30$\times$ by the end of training. A padding of 20 grids is applied to all FNOs. 
%An optional warm-up step could be added that we found didn't influence model performance.

We consider only the material subdomain model in this analysis for simplicity and specify architectures with similar floating point operations (FLOP) and model weights. All data in Table \ref{sub-bench-table} and plots in Figure \ref{fig-3} are conducted on 10k newly generated, unseen test data. We see that the specification of a targeted model architecture is crucial to achieving high accuracy. The vanilla FNO fails to learn the problem with good accuracy, even with a large number of model weights. While the Swin transformer requires relatively fewer neural network weights, the expensive self-attention operations require over 10x FLOPs compared to FNO-based architectures.  A comparison of SM-FNO-data-only and SM-FNO indicates that the explicit inclusion of Maxwell's equations leads to a dramatic reduction of $L_{pde}$ and $L_{bc}$, which is essential to getting the DDM algorithm to converge. Our largest model, SM-FNO-v2, is 99.0\% accurate and features exceptionally low $L_{pde}$ and $L_{bc}$. 

\begin{figure*}[!ht]
  \centering
  \makebox[\textwidth][c]{\includegraphics[scale=0.47]{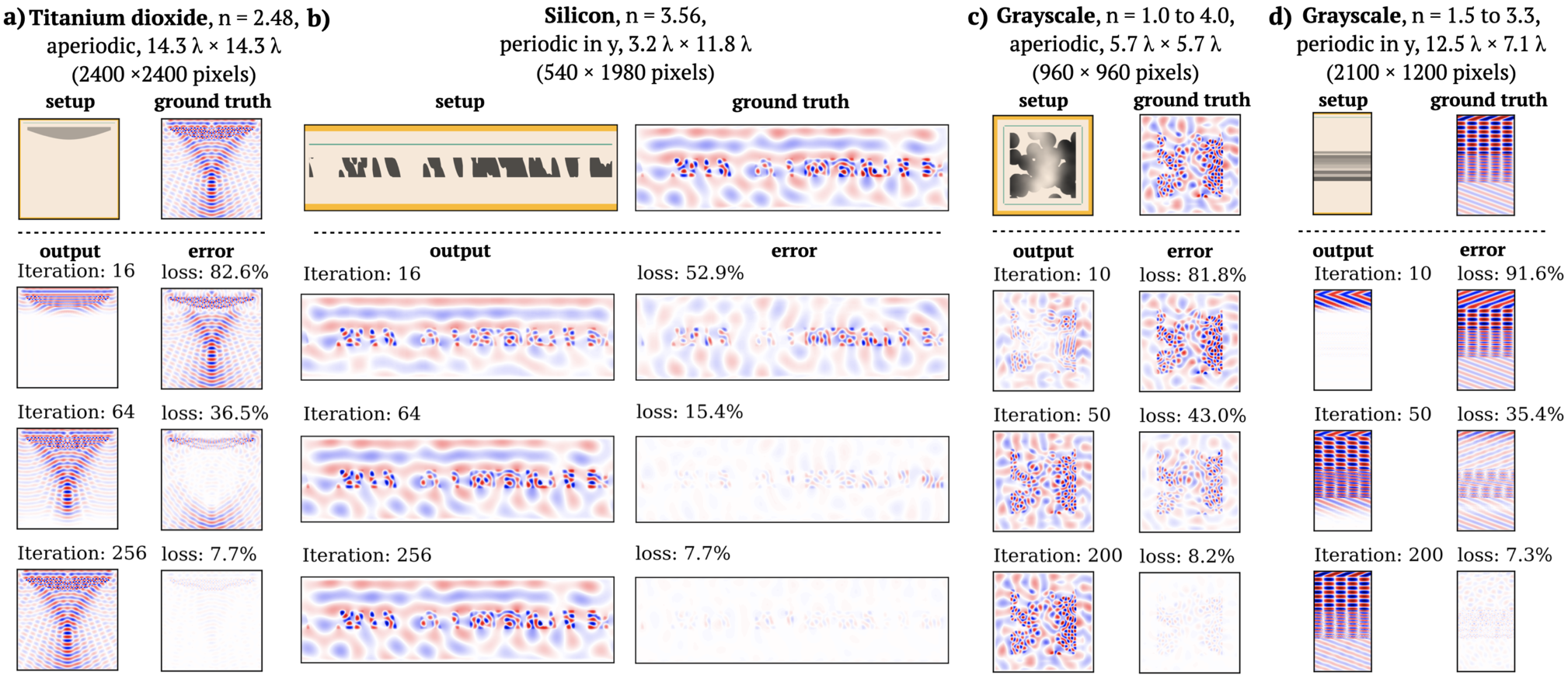}}%
  \caption{SNAP-DDM evaluated on different electromagnetics systems.  These systems include: \textbf{a)} a titanium dioxide microlens, \textbf{b)} a thin film silicon-based metasurface, \textbf{c)} a volumetric grayscale metamaterial scatterer, and \textbf{d)} an optimized grayscale thin film stack featuring high reflectivity. The simulation domain contains grid resolution with physical dimensions of $6.25 nm$ and the wavelength $\lambda=1.05 \mu m$. Ground truth fields, model output fields, and field error are plotted on the same scale for each device. The largest simulation domain is in (a) and comprises an array of $40 \times 40$ subdomains. }\label{fig-4}
\end{figure*}

Model FLOPs are computed using the open-source library \textit{fvcore}. The FLOPs of FFT operations are computed using the formula: $2 L (N C^2 + N C \log N)$ for 2d FFTs, and $2 L ((H+W) C^2 + N C (\log H + \log W))$ for 1d FFTs, in which $N=HW$ is the number of grids of each channel, $L$ is number of layers and $C$ is number of channels \cite{guibas2021adaptive}. The factor $2$ accounts for forward and inverse FFT operations.
% \begin{table}[!ht]
%       \caption{Subdomain model benchmark}
%       \label{sub-bench-table}
%       \begin{center}
%       \begin{tabular}{p{3cm}|ccc|ccc|p{1cm}p{1cm}}
%         \Xhline{4\arrayrulewidth}
%         % \multicolumn{2}{c}{Part}                   \\
%         \multirow{3}{*}{Model}  &    \multicolumn{3}{c|}{Electromagnetics} &  \multicolumn{3}{c|}{Fluid flow} &  \multirow{3}{0.8cm}{Param (M)} &  \multirow{3}{0.8cm}{FLOP (G)}     \TBstrut\\ 
%         & $L_{data}$ & $L_{pde}$   & $L_{bc}$        & $L_{data}$  & $L_{pde}$  & $L_{bc}$ &  & \\
%         & (\%)       & (a.u.)      & (a.u.)          & (\%)        & (a.u.)     & (a.u.)   &  &      \\
%         \hline
%         FNO         & 38.7  & 1.77  & 0.30  & 6.46 & 2.44 & 0.34 & 69.2  & 0.78 \Tstrut\\
%         UNet        &  5.4  & 0.73  & 0.10  & & & & 5.2   & 1.53 \\
%         Swin T      &  5.2  & 2.15  & 0.14  & & & & \textbf{1.9}   & 9.60\\
%         SM-FNO(ours)&  \textbf{3.9}  & \textbf{0.50}  & \textbf{0.06} & 1.92 & 0.21&  0.07 & 4.7   & \textbf{0.66}  \Bstrut\\
%         \hline
%         SM-FNO-large(ours) & \textbf{1.0} & \textbf{0.30} & \textbf{0.03} & & &  & 10.1 & 1.43
%         \TBstrut\\
%         \Xhline{4\arrayrulewidth}
%       \end{tabular}
%       \end{center}
%     \end{table}

\subsection{Large scale electromagnetics simulations}
Large scale electromagnetic simulations  comprising high contrast heterogeneous media are notoriously hard to solve using end-to-end neural surrogate solvers. We show in Appendix \ref{appendix:end-to-end-FNO} that the training of a Fourier neural operator to solve full-scale problems leads to fundamental scaling bottlenecks in dataset size, model size, and memory usage. We also show in Appendix \ref{appendix:end-to-end-PINN} that PINNs struggle to scale up to large simulation domains comprising high dielectric contrast media, and that the solutions produced from trained PINN models are particularly sensitive to their detailed initialization and training conditions.  These results are consistent with recent large scale simulation demonstrations in the literature: one concept based on   
graph networks featured errors of 28\% \cite{khoram2022graph} and another concept based on neural operators featured errors ranging from 12\% to 38\%  \cite{gu2022neurolight}. 

On the other hand, our SNAP-DDM algorithm combining  trained subdomain surrogate solvers with the overlapping Schwartz DDM method produces a qualitatively different and better result.
To demonstrate, we solve a variety of large scale electromagnetics problems featuring a wide range of dielectric constant and domain size configurations.  We use the SM-FNO-v2 architecture for the material and PML models and the lighter SM-FNO-v1 network for the source model. For each problem, the global domain is initially subdivided into an array of subdomains, each of which are  classified as PML, source, or material subdomains.  During each SNAP-DDM iteration, data from subdomains of a given class are aggregated into a batch and inputted into the corresponding specialized SM-FNO, which infers and outputs the H-field solutions of the batch in a parallelized manner.  The DDM algorithm is stopped after a predetermined number of iterations.
%The simulated subdomains then update their boundaries conditions for next iteration until the total field error reaches some threshold. 
%Each class of subdomains are bundles in the batch dimension and processed by the corresponding specialized subdomain model in parallel. 
%Iterative methods, on the other hand, offer unique advantage and flexibility. We utilize our trained subdomain surrogate solvers with the overlapping Schwartz DDM method to perform large domain electromagnetic simulations. 
%Our scheme can adapt to global domains of arbitrary size by tiling different numbers of subdomains together and tailoring the amount of overlap between subdomains.

Representative electromagnetic simulation results are shown in Figure \ref{fig-4} and demonstrate the versatility and accuracy of SNAP-DDM.  The simulations feature widely varying global domain sizes, indicating the ability for our scheme to readily adapt to arbitrary global domain sizes through the tiling of different numbers of subdomains and tailoring the amount of overlap between subdomains.  Some of these  simulations feature the use of PML boundaries on all sides, which is ideal for purely scattering simulations, while others comprise half PML and half Bloch periodic boundaries, which are a natural boundary choice for semi-periodic systems. 
The off-normal incident field in the thin film problem is achieved by tailoring the line source profile with the appropriate Bloch phase.  For all of these examples, the final and ground truth fields appear indistinguishable, and the absolute error in the final fields in all cases is near 5\%.  

\begin{figure}[!hb]
      \centering
      \includegraphics[scale=0.25]{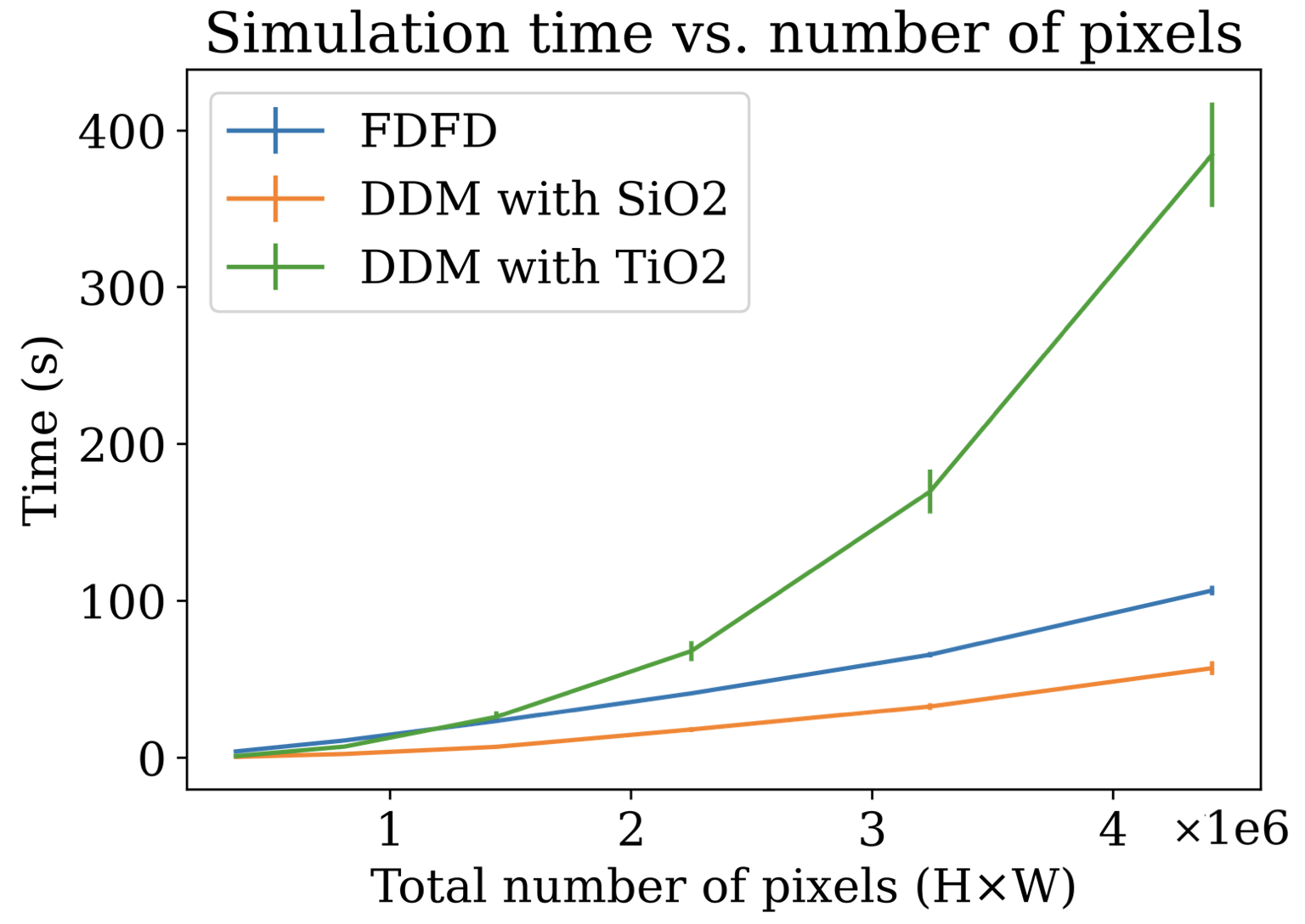}%
      \caption{Time complexity comparison with SNAP-DDM and a conventional FDFD solver for two different dielectric materials. 
       }\label{fig-5}
\end{figure}

\begin{table*}[!hb]
      \caption{Steady state flow: subdomain model benchmark on 10k test data}
      \label{sub-bench-table-flow}
      \begin{center}
      \begin{tabular}{llllll}
        \toprule
        Model      & $L_{data}$ (\%) & $L_{pde}$ (a.u.)  & $L_{bc}$ (a.u.)      & Param (M) & FLOP (G)     \\ 
        \midrule
        FNO         & 6.5 & 1.77  & 2.44  & 69.2  & 0.34 \\
        Swin T      &  5.1  & 1.10  & 0.13  & \textbf{1.9}   & 9.60\\
        UNet        & 1.8  & 0.35  & 0.10  & 5.2   & 1.53 \\
        SM-FNO(ours)&  \textbf{1.3}  & \textbf{0.23}  & \textbf{0.06}  & 4.7   & \textbf{0.66}  \\
        \bottomrule
      \end{tabular}
      \end{center}
    \end{table*}
\begin{figure*}[!hb]
      \centering
      \makebox[\textwidth][c]{\includegraphics[scale=0.47]{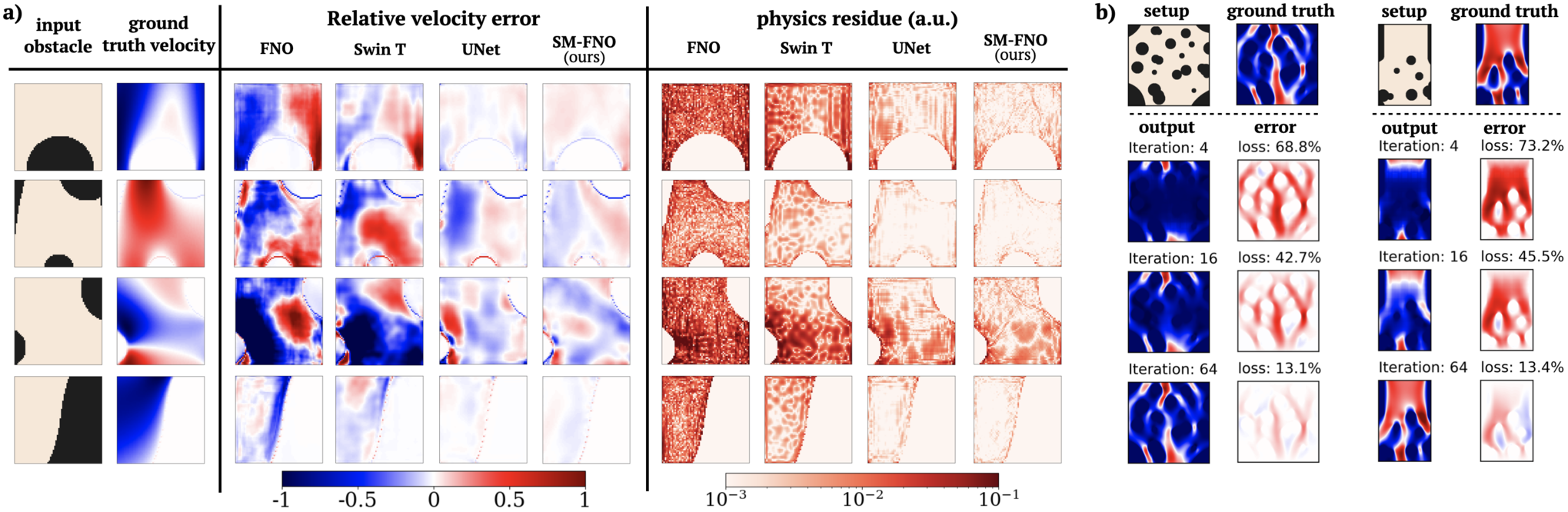}}%
      \caption{SNAP-DDM pipeline for steady-state fluid flow problems. \textbf{a)} Subdomain model benchmark. \textbf{b)} Steady state fluid flow velocity fields solved with SNAP-DDM. Subdomain grid sizes are $8 \times 8$ (left) and $8 \times 5$ (right).} \label{fig-6}
\end{figure*}

\subsection{Time complexity}
In this section, we benchmark the time complexity between SNAP-DDM and a conventional FDFD solver for 2D EM problems. Based on the current accuracy level of SNAP-DDM, we choose to benchmark the time it takes to reach an average mean absolute accuracy of 15\%, evaluated on 10 random devices for each domain size. For SNAP-DDM, the number of iterations required for convergence strongly depends on the material refractive index, and we therefore benchmark computation time for simulation domains containing either  silicon dioxide (n = 1.5) or titanium dioxide (n = 2.48).  Square domains with sizes ranging from $600 \times 600$ grids to $2100 \times 2100$ grids are evaluated. The FDFD benchmark simulations are performed on 10 grayscale dielectric structures for each domain size with refractive indices ranging from those of silicon dioxide to titanium dioxide.  SNAP-DDM is run with one NVIDIA RTX A6000 GPU, and the FDFD solver runs on single CPU of model Intel Xeon Gold 6242R. From the time benchmark, we observe that for silicon dioxide, which has a relatively low refractive index, SNAP-DDM has better performance compared to the FDFD solver.  However, for titanium dioxide, SNAP-DDM requires significantly more iterations and ultimately takes a much longer time than FDFD. Many aspects of SNAP-DDM can be optimized to improve its efficiency, and we show in Appendix \ref{appendix:subdomain-size} that by doubling the subdomain size from 64 by 64 grids to 128 by 128 grids, the overall DDM algorithm becomes twice as efficient.

\subsection{SNAP-DDM steady state fluid flow simulations}
%The aforementioned data generation pipeline, subdomain model training and DDM algorithm construction could be applied to other PDE systems as well. Here we make the attempt to simulate 2d steady state fluid flow problems, 
The SNAP-DDM concept can apply to a broad range of steady state PDE problems, and we demonstrate here the application of SNAP-DDM to 2D steady state fluid flow problems.  Fluid mechanics systems are governed by the incompressible Navier-Stokes (NS) equation, which is:
\begin{equation}\label{NS-equation}
\frac{\partial \bm{u}}{\partial t} + (\bm{u} \cdot \nabla)\bm{u} - \nu \nabla^2\bm{u} = -\frac{1}{\rho}\nabla p
\end{equation}
We solve steady state flows in an arbitrary-shaped pipe with  circularly shaped obstacles and a viscosity of $\nu=0.08$ \cite{chen1998lattice}. To train our subdomain boundary value solvers for these problems, we first simulate flows using the time domain Lattice-Boltzmann Method and run the simulations until the flows are at steady state. A total of 200 ground truth simulations with $900 \times 600$ grids are generated, from which the data for 100k subdomains with $64 \times 64$ grids is produced as the subdomain training dataset.  Further details are in Appendix \ref{appendix:data-generation}. The subdomain model takes an image of the obstacle and velocity field ($u, v$) Robin boundary conditions as inputs, and it outputs images of the full velocity field.  Ground truth pressures are used with the steady state version of Equation (\ref{NS-equation}) to compute physics loss.

\begin{table*}[!ht]
  \caption{Ablation study}
  \label{ablation-table}
  \centering
  \begin{tabular}{lllllllll}
    \toprule
    \multicolumn{1}{c}{Model}       & $L_{data}$ & $L_{pde}$  & $L_{bc}$ & Param & FLOP & L & C & M  \\
     & (\%) & (a.u.) & (a.u.) & (M) & (G) & & &\\
    \midrule
    FNO without residual connection & 16.01 & 1.92  & 0.217  & 41.0  & 0.97 & 4 & 100 & 16 \\
    SM-FNO remove residual connection    & 7.61   & 1.03  &  0.123 & 8.9  & 0.78  & 4 & 44 & 16\\
    SM-FNO remove self modulation - 1 & 8.04 & 1.93  & 0.166  & 42.0  & 1.00 & 10 & 64 & 16\\
    SM-FNO remove self modulation - 2 & 6.11 & 3.29  & 0.155  & 32.8  & 0.82 & 20 & 40 & 16\\
    SM-FNO-v1&  3.85 &  0.50 & 0.067  & 4.7  & 0.66 & 16 & 16 & 16\\
    SM-FNO-v2-data-only & 1.36   & 2.76  &  0.073  & 10.2 & 1.43 & 16 & 24 & 16\\
    SM-FNO-v2 &  1.00 &  0.30 & 0.030 & 10.2 & 1.43 & 16 & 24 & 16\\
    \bottomrule
  \end{tabular}
\end{table*}

\begin{figure*}[!hb]
      \centering
      \makebox[\textwidth][c]{\includegraphics[scale=0.5]{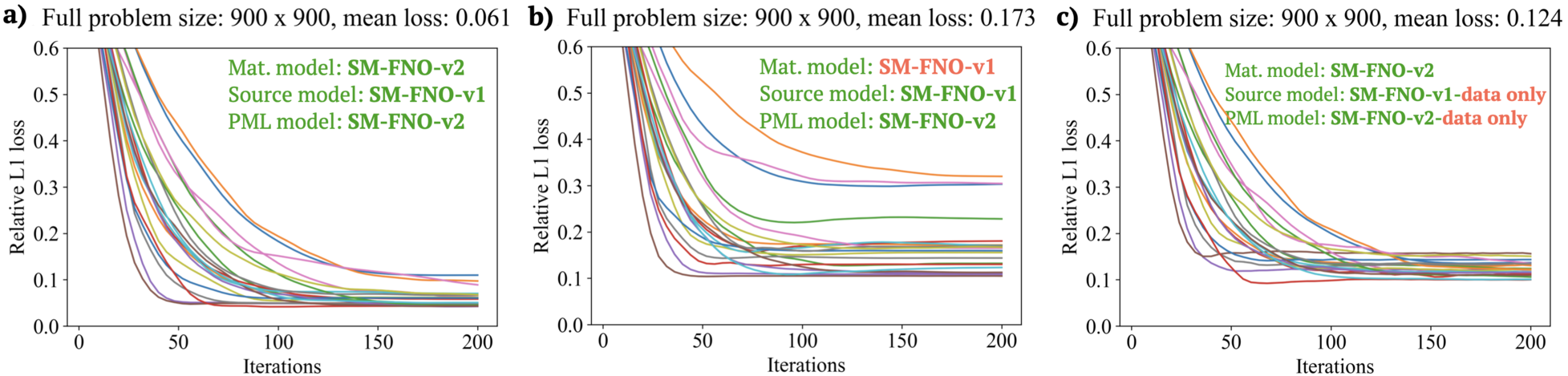}}%
      \caption{SNAP-DDM convergence curves with different types of subdomain solvers.  The plots show DDM algorithm error versus iteration count, with each curve representing field error from a simulated random grayscale device within a $900 \times 900$ grids domain (15 by 15 subdomains).  \textbf{a)}: The proposed setup with 3 specialized subdomain models. \textbf{b)} Same as (a) but use of a lighter material model trained on 100k data. \textbf{c)} Same as (a) but use of source and PML models trained using only data loss. The slower convergence curves correspond to materials with higher average dielectric constant. }\label{fig-7}
\end{figure*}

Benchmark results of our SM-FNO subdomain solver are summarized in Table \ref{sub-bench-table-flow} and Figure \ref{fig-6}a, and they indicate that our SM-FNO network displays the lowest data and physics loss compared to alternative subdomain solver architectures, with data loss approaching 1\%.  Demonstrations of steady state fluid flow simulations with SNAP-DDM are shown in Figure \ref{fig-6}b, where the simulations produce velocity profiles with errors less than 15\%.  The reduced accuracy in the flow SNAP-DDM simulations, compared to those from electromagnetics, is likely due to the sub-optimal performance of Schwartz DDM with Robin boundary conditions for steady state flow problems.  DDM for fluids problems continues to be a topic of active research, and further improvement in SNAP-DDM for fluids will be followed up in future work.

%our  solves for  are the benchmark of subdomain models for fluid flow problems, and boundary value problem demonstrations that are solved using the SNAP-DDM framework with . While we obtain subdomain models achieving similar near-unity accuracy, the Schwarz method appears to be not as stable and accurate compared to electromagnetic setups. We believe performance could be improved by changing Dirichlet boundary condition to transparent boundary conditions proposed in Optimized Schwarz Methods tailored for NS equations\cite{tourrette2001absorbing, bruneau2001towards, blayo2016towards}, which we leave it to be explored in follow-up works.
    
\section{Ablation study}\label{ablation}
%\subsection{Subdomain model architecture}
We perform an ablation study to understand the contribution of each modification to the FNO featured in our SM-FNO. Results are shown in Table \ref{ablation-table} where L is the number of layers, C is the number of channels (hidden dimension), and M is the number of Fourier modes for linear transform. We start with the vanilla FNO without residual connections and find that the model fails to consistently learn for depths larger than 4 layers. Upon training with 4 layers, increases in hidden dimension and number of modes increases model size without contributing  much to performance. When self modulation is removed from the SM-FNO, deeper networks could be trained with the residual connection but different depths and widths produced similar sub-optimal performance. This indicates that without the modulation path, model expressivity is limited.  When we remove the residual connections from the SM-FNO, the model did not work well with large depths. The best model contained 4 layers and produced reasonable accuracy. 

It is clear the two modifications that we added to the FNO architecture are both synergistic and essential to improving subdomain solver accuracy: the residual connection enables deep architectures to be trained while the self-modulation connection increases the model expressivity by promoting self-multiplicative interactions within each input. In addition, the hybrid physics-augmented training scheme significantly lowers the physical residue while slightly reducing the data loss.  We also point out that our use of Robin boundary conditions ensures that the subdomain solvers solve a well-posed PDE problem, unlike alternative boundary conditions such as Dirichlet boundary conditions.

%\subsection{DDM algorithm accuracy} \label{sec:DDM-ablation}
The accuracy of the SNAP-DDM framework is dependent on multiple factors. Plots of DDM accuracy versus iterations for 20 devices under different setups is shown in Figure \ref{fig-7}. These plots show that error from SNAP-DDM increases significantly when we replace the large material model with a lighter version, indicating the need for the subdomain models to have near unity accuracy.  These plots also show that when a fraction of models is trained without physics, SNAP-DDM error also increases, indicating the need for hybrid data-physics training for all subdomain models.

\section{Limitations and future work}
There are multiple potential speedup strategies with SNAP-DDM that can be considered in future work.  Preconditioning is a common strategy in DDM for improving converg speed by reducing the condition number of the system, especially for large ill-conditioned problems  \cite{vion2014double, gander2019class}. Increasing the subdomain size, and more generally incorporating a non-uniform grid or mesh-to-grid approach to subdomain solving, has the potential to introduce computational savings\cite{liu2023nuno}, though a further quantitative analysis into the balance between model size and latency is required. More sophisticated SNAP-DDM implementations may incorporate multi-level or multi-grid concepts for initialization and improved memory management, as well as higher order boundary conditions. 
% We also anticipate that higher order boundary conditions can help with DDM convergence and accuracy. 

\section*{Acknowledgement}
This work is funded by National Science Foundation under Award Number 2103301, Office of Naval Research under Award Number N00014-20-1-2105, and 3M under Award Number IC2020-2735.

\section*{Impact Statement}
This paper presents work whose goal is to advance the field of 
Machine Learning. There are many potential societal consequences 
of our work, none which we feel must be specifically highlighted here.

\clearpage
\bibliography{submission}
\bibliographystyle{icml2024}

\newpage
\appendix
\onecolumn
Code for this project could be found at: 

\url{https://github.com/ChenkaiMao97/SNAP-DDM}
\section{Reference for DDM methods}
\label{appendix:DDM-reference}
The Additive Schwarz Method (ASM) is a simple yet powerful domain decomposition method that is naturally suited for the parallel computation of PDE problems posed using both finite difference and finite element methods. The convergence of the ASM has been theoretically proven for positive-definite systems with elliptical PDEs\cite{dolean2015introduction}, and it can be accelerated using preconditioners as in the Restricted Additive Schwarz (RAS) method \cite{cai1999restricted}, which is used in scientific computation software like PETSc \cite{balay1997efficient}. For indefinite systems like Maxwell’s Equations, the use of Dirichlet boundary conditions leads to divergence, and Optimized Schwarz Methods (OSM) are required that utilize robin type boundary conditions \cite{gander2006optimized} or higher order and non-local boundary conditions \cite{gander2002optimized} to improve convergence.

Beyond the Schwarz Method, two-level and multi-grid methods speed up the convergence by exchanging global information using a global coarse space on top of the subdomain solvers. An example involving the Helmholtz equation is the construction of the Dirichlet-to-Neumann coarse space \cite{conen2014coarse}, which inherently applies to heterogeneous problems. Other popular methods include Neumann-to-Neumann and Finite element tearing and interconnecting (FETI) methods \cite{dolean2015introduction,farhat1991method}, which construct a reduced interface system and use Lagrange multipliers to ensure a weak continuity of the solution across the interfaces.

While the current study only uses Neural Network surrogate solvers in the optimized Schwarz Method setup, we believe the efficient solving of general subdomain problems is central to accelerating all classes of domain decomposition methods.
\section{Loss functions}
\label{appendix:loss-function}
Here we present the data-physics loss function used to train the subdomain models:
\begin{equation}\label{dataloss}
    L_{data} = \frac{1}{N}\sum_{n=1}^{N}  \Big|\Big|\mathbf{H}^{(n)}-\hat{\mathbf{H}}^{(n)}\Big|\Big|_1 
\end{equation}
\begin{equation}\label{PDEloss}
    L_{pde} = \frac{1}{N}\sum_{n=1}^{N}\Big|\Big|\nabla \times (\frac{1}{\varepsilon(\mathbf{r})} \nabla \times \mathbf{H}^{(n)}) - \mu_0 \omega^2 \mathbf{H}^{(n)}\Big|\Big|_1  
\end{equation}
\begin{equation}\label{PDEbc}
    L_{bc} = \frac{1}{N}\sum_{n=1}^{N}\Big|\Big|g(\mathbf{r}) - \big(jk(\mathbf{r})\mathbf{H}^{(n)} -\frac{\partial \mathbf{H}^{(n)}}{\partial n}\big)\Big|\Big|_1 
\end{equation}
in which $\hat{\mathbf{H}}$ is the ground-truth magnetic field, $g(\mathbf{r})$ is the input boundary value, $k(\mathbf{r}) = 2\pi \varepsilon(\mathbf{r})/\lambda$ is the wave vector in the medium. 
\section{Data generation} \label{appendix:data-generation}
To generate random geometric distributions for 2D EM problems that include both regular-shaped objects and structures with free-form topologies, we adopt a pipeline inspired from image processing:
\begin{enumerate}
    \item Generate random noise map in range $(0,1)$.
    \item Threshold by some level between $0$ and $1$.
    \item Erode the map with a Gaussian filter.
    \item Dilate the map using filters with tilted elliptic profiles.
    \item Apply Gaussian filter for smoothing.
\end{enumerate}
By tuning the threshold level, filter weight, and erosion and dilation parameters, random shaped geometries with different feature size distributions are generated. We generate Gaussian random fields and Voronoi diagrams with grayscale values between $1$ and $16$, and then we use the random geometries as masks to create the grayscale material distributions used to produce ground truth data. The Gaussian random field creates a continuously changing dielectric that are representative of features appearing in freeform metamaterial designs. The Voronoi diagram creates boundaries between different constant regions that produces material boundary features in the training data. 

Line sources with a mix of random sinusoidal profiles are placed on all four sides of the generated grayscale material to create randomly scattering fields in all directions. Uniaxial PML boundaries are places on all four sides with thickness of 40 grid cells. We use ceviche FDFD solver to generate 1000 fullwave simulations of size 960 by 960 grids, with grid resolution of 6.25nm and wavelength of 1050nm. We then cropped the fields and physical properties to produce an 100k material dataset, an 1M material dataset, an 100k source dataset and an 100k PML dataset for training specialized subdoamin models.

\begin{figure}[!ht]
      \centering
      \makebox[\textwidth][c]{\includegraphics[scale=0.5]{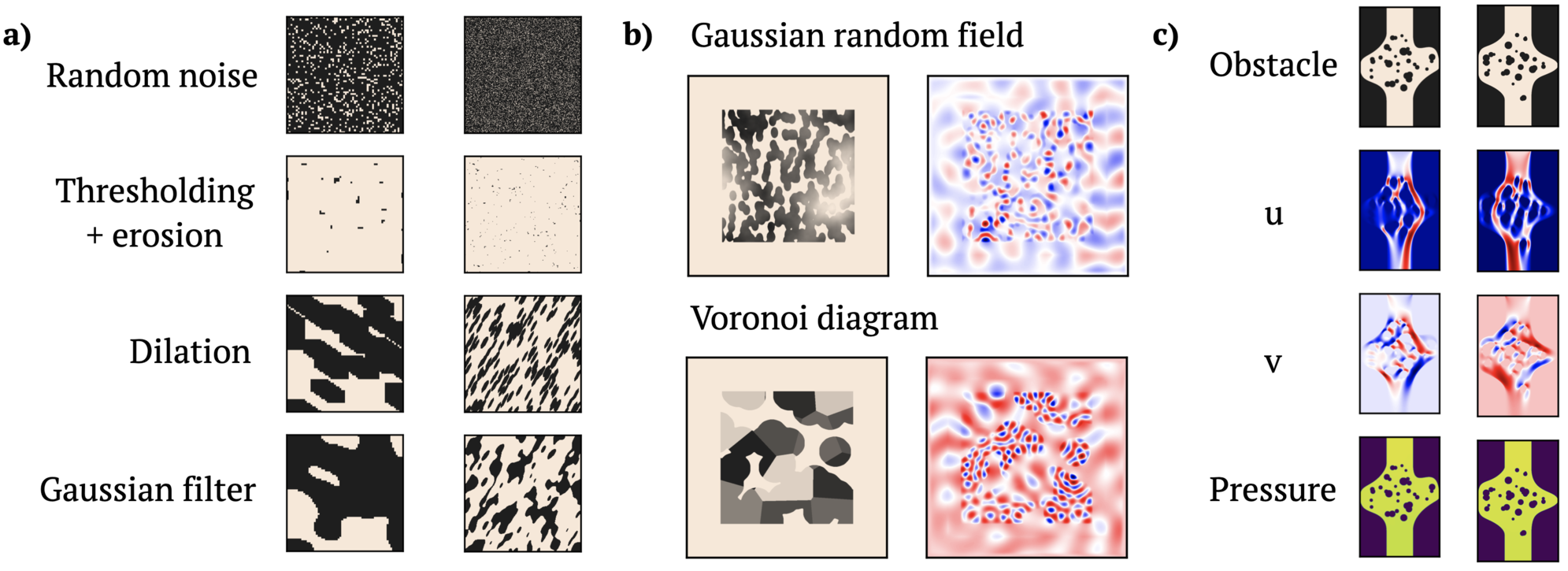}}%
      \caption{Pipeline for data generation. \textbf{a)} Image processing-inspired procedure for generating random dielectric geometries. \textbf{b)} Two kinds of grayscale geometries and corresponding simulated H-fields. \textbf{c)} Flows in arbitrarily-shaped pipe with circular obstacles.} \label{fig-8}
\end{figure}

For steady state fluids simulations, we generate pipes with arbitrarily-shaped middle sections that are created by connecting two random Bézier curves. The idea is to create boundaries beyond straight vertical walls that may appear in freeform flow flow scenarios. Circular obstacles with randomly sampled radii are placed in the pipe in a manner where a minimum gap size is guaranteed. We use constant velocity($u=u_0$, $v=0$) as the boundary condition for the  inlet surface. For the bottom outlet, $v=0$ and $P=constant$ is used as the boundary condition. The viscosity is fixed to be $0.08$ and steady state flow is reached when the relative velocity change after 100 time steps is less than $10^{-4}$. The steady state solution is not guaranteed in this way, but we found that 199 out of 200 simulations reached steady state.  For  both EM and fluids cases, subdomain data is produced by cropping data from large-scale simulations with optional rotation as data augmentation method.
%(alternatively a Poiseuille flow velocity profile could be given)

\section{Hybrid data-physics training scheme} \label{appendix:physics-training}

The subdomain solvers are trained using a hybrid loss function composed of a data term, $L_{data}$, and a physics loss term, $L_{physics}$, which is scaled by a hyperparameter $\alpha$:

$$ L = L_{data} + \alpha \cdot L_{physics}. $$

Previous work in physics-augmented neural network training has demonstrated that training convergence and performance is sensitive to the \textit{relative} magnitude of the physics loss term compared to the data loss term \cite{chen2022high}. To maximize the generality of the proposed training setup, we employ a dynamically tuned hyperparameter, $\alpha$, which is scaled throughout the training process, like the approach taken in the WaveY-Net study \cite{chen2022high}. At the end of each epoch, $\alpha$ is modified such that the ratio between $\alpha \cdot L_{physics}$ and $L_{data}$ is a constant, $\alpha'$, throughout the entire training process. The practice of dynamically tuning the physics loss coefficient greatly stabilizes training convergence across different simulation problems, thereby allowing a working training scheme to readily generalize to problems governed by different physics equations. 

The constant physics ratio, $\alpha'$, is neural network model dependent and appears insensitive to the two types of problems being simulated. All the FNO models are trained with $\alpha'=0.3$ and the U-Net and Swin Transformer are trained with $\alpha'=0.1$. However, due to the slower learning rate of vision transformers compared to convolutional neural networks, $\alpha'$ is set to $0$ for the first 50 training epochs to prevent divergent training behavior. Divergent behavior occurs in domains with high contrast material due to the presence of strong optical resonances. Without substantial influence from data to push the optimization in regimes with the correct resonances, unstable interplay between bulk and boundary physics loss leads the optimization process astray.

\section{Implementation details for U-Net} \label{appendix:baseline-unet}
The U-Net architecture employed in this study is constructed as follows:

\begin{itemize}
    \item Each half of the U-shaped architecture contains 5 blocks of convolutional layers;
    \item Each convolutional block is composed of 6 convolutional layers;
    \item Each convolutional layer is composed of a sequence of operations: convolution, batch normalization, and ReLU activation;
    \item The number of convolutional kernels contained in the convolutional layers of each block is increased by a factor of two compared to the previous block's number, starting with 30: $30\cdot(2^{(block-1)})$. The number in the blocks of the second half of the U-shaped architecture mirror those in the first half.
\end{itemize}

\section{Implementation details for Swin Transformer} \label{appendix:baseline-swint}
We designed our implementation of the Swin Transformer  with the same shifting windows and windowed attention as featured in previous architectures \cite{liu2021swin}. Notably, we abstain from utilizing patch merging, as our network design maintains consistent input and output dimensions. Our architecture comprises multiple stacked Swin Transformer layers, all configured with uniform patch sizes. The hyperparameters regarding the model is chosen based on the Swin transformer model used in image semantic segmentation tasks, and adjusted based on the input size of our subdomain problem.

To elaborate on the architectural parameters:

\begin{itemize}
    \item Patch Size: We employ a patch size of 1.
    \item Window Size: Each attention window spans 9 patches.
    \item Number of Heads: Multi-head attention is applied with 16 attention heads.
    \item Swin Transformer Blocks: Each layer of the network contains 16 Swin Transformer blocks with window shifting enabled.
    \item Layers: The network is formed by stacking 4 such layers.
\end{itemize}

We initialize trainable absolute positional encodings drawn from a normal distribution with a mean of 0 and a standard deviation of 0.01. For encoding domain-specific information, such as the input refractive index in EM simulations or obstacles in fluid simulations, we employ a matrix multiplication to transform these data into a 48-dimensional vector. Boundary conditions are treated similarly but encoded using a separate encoder. These encodings are then padded around the original image for 4 times, leading to an overall 72 by 72 input to the network. Corners not covered are left as 0.  The output of the network is subsequently transformed using another matrix multiplication to ensure it conforms to the dimensions required for the final output. For training the model, we use Adam optimizer with learning rate set to 0.001 for initial, and then exponentially decay it to 0.0001 over the training course of 50 epochs. We do not use dropout or weight decay during training.

\section{Impact of different subdomain sizes}
\label{appendix:subdomain-size}
As pointed out in the main text, the runtime optimization of SNAP-DDM-like systems is a broad topic to be explored in future works. Here we present an immediate runtime improvement by increasing the subdomain size from 64 grids to 128 grids (with the same grid resolution). 

We train an 128 by 128 subdomain model with the same SM-FNO-V2 architecture, which in this case has 11.3M weights. We crop the same large simulation data to produce a 1M training dataset of 128 by 128 grids subdomains. The final loss on the material model is 2.06\%, about twice as much as our 64 by 64 model, which is expected for the increased subdomain domain size.

We then run the DDM benchmarks shown in the table below, in which we choose 4 different problem sizes, and run DDM algorithm with both 64 and 128 sized models. The iteration and run time are recorded when the averaged MAE loss drops below 10\% for 10 random problems with TiO2 as material.

\begin{table*}[!ht]
  \caption{DDM benchmark with \textbf{64 by 64} grids subdomain model}
  \begin{center}
  \begin{tabular}{llll}
    \toprule
    Problem size & Subdomains & Iterations  & Total time (s)    \\ 
    \midrule
    604 by 604  & 100 & 37 & $2.052\pm0.006$ \\
    968 by 968	& 256 & 77 & $11.623\pm0.114$ \\
    1208 by 1208 & 400 & 111 & $26.788\pm0.164$ \\
    1444 by 1444 & 576 & 153 & $53.734\pm0.153$ \\
    \bottomrule
  \end{tabular}
  \end{center}
\end{table*}

\begin{table*}[!ht]
  \caption{DDM benchmark with \textbf{128 by 128} grids subdomain model}
  \begin{center}
  \begin{tabular}{llll}
    \toprule
    Problem size & Subdomains & Iterations  & Total time (s)    \\ 
    \midrule
    604 by 604  & 25 & 18 & $0.900\pm0.001$ \\
    968 by 968	& 64 & 38 & $5.252\pm0.029$ \\
    1208 by 1208 & 100 & 52 & $11.526\pm0.064$ \\
    1472 by 1472 & 144 & 76 & $24.771\pm0.065$ \\
    \bottomrule
  \end{tabular}
  \end{center}
\end{table*}

The experiment shows that with the same amount of ground truth data generated, when we scale subdomain model size from 64 to 128, with the same architecture and both trained with 1M data, the subdomain model sacrifice in accuracy, but the DDM algorithm gets twice as fast. An interesting optimization problem arises in balancing subdomain size and model size to achieve the best overall efficiency.

\section{End-to-end FNO on the full problem}
\label{appendix:end-to-end-FNO}
In this section, we demonstrate the results and difficulties in training an SM-FNO model on the full-sized problem consisting of 960 by 960 grids. The training data consists of a total of 1000 data samples, which is split into 900 training samples and 100 test samples. Each input device consist of a random grayscale dielectric material map, a map of 4 line sources with a random profile, and a PML map produced by a 40-grid thick UPML on four sides, which is constant.

We trained an SM-FNO model with 6 layers, 64 channels and 16 Fourier modes. The model has 262M weights and 531G FLOP per input device. The Adam optimizer is  used with learning rate starting at 3e-4 and annealed to 1e-5 over 100 epochs. We used an NVIDIA RTX A6000 GPU with 48GB memory, and could fit up to a batch size of 4 during training.

\begin{figure*}[!ht]
      \centering
      \makebox[\textwidth][c]{\includegraphics[scale=0.47]{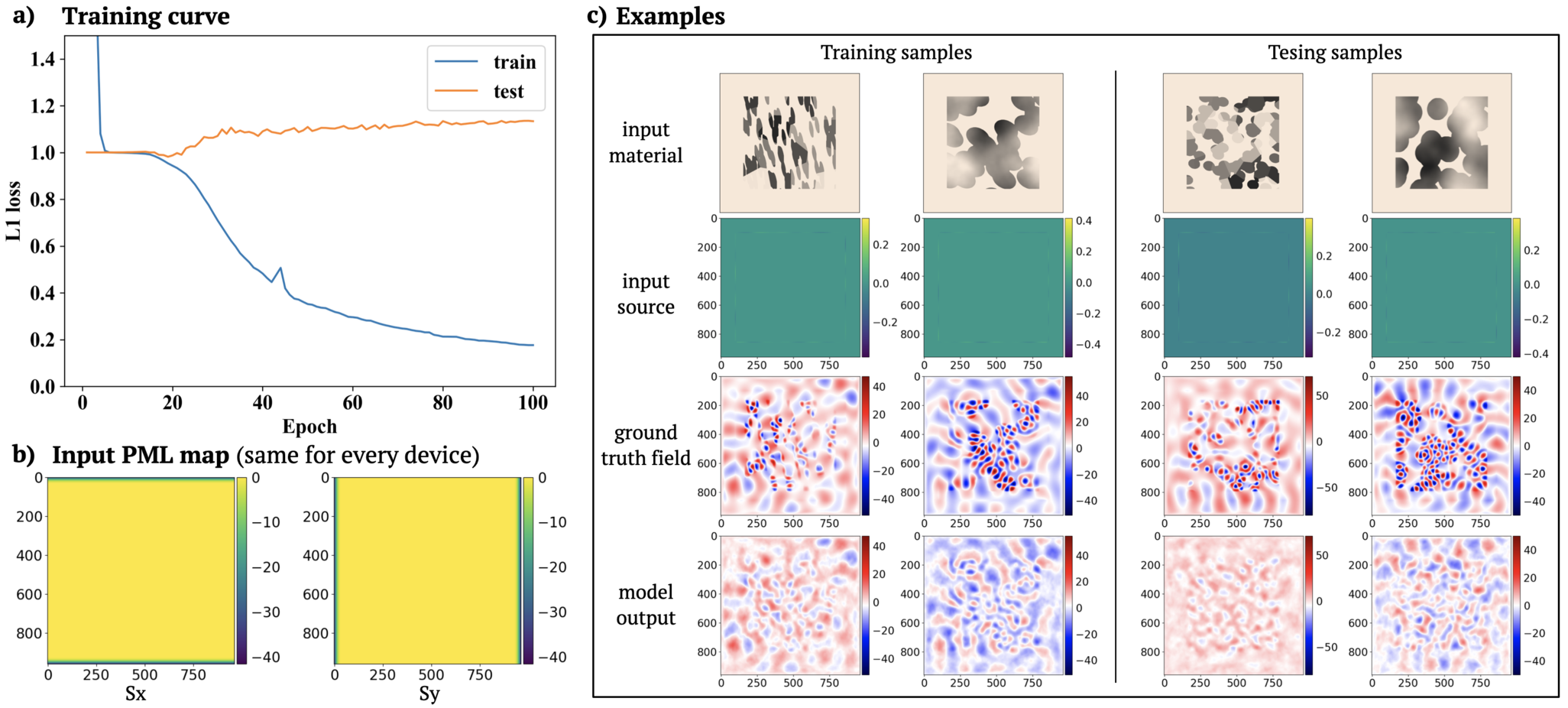}}%
      \caption{SM-FNO trained on full-sized problem with 960 by 960 grids. a) Training curve for 100 epochs. b) The input PML map shared for each device. c) Samples from the training and testing datasets. 
       }\label{SI-S2}
\end{figure*}

From the training curve and sample visualizations, it is clear that the model is able to overfit to the training data and learn the lower frequency spectrum of the fields, but fails to generalize to test data. This is expected as we have seen that we need more than 100k data even for a 64 by 64 grids subdomain with heterogeneous material and arbitrary boundary. It is expected that orders of magnitude more data is required to learn similar problems on a larger scale. 

We believe it is theoretically possible that with sufficient resources and time, an end-to-end model could be trained for a large problem. As a quick comparison, it takes about 2 hours to generate the 1000 training samples on a desktop with 40 CPU cores. It would take over 2 months to generate 100k data. The scaling would be even worse for problems in 3D. 

At the same time, we have demonstrated that cropping the same 1000 simulations to create a subdomain dataset with size on the order of 100k to 1M could be sufficient for building semi-general subdomain solvers that are capable of solving a broad range of problems.

Other methods could be beneficial in practice, like generating low-resolution data and using physics to help formulate high-resolution solutions \citep{li2021physics}. We note that for wave-like problems, a minimum number of points needs to be sampled per wavelength to avoid aliasing, which sets the lower bound for problem complexity. Besides, end-to-end methods are usually designed for a fixed physical domain size, while the DDM approaches  have the flexibility to be applied to different sized problems.

\section{End-to-end PINN on the full problem}
\label{appendix:end-to-end-PINN}
In this section, we consider a broader comparison with physics-informed baselines by evaluating the performance of PINNs on the same simulation setup illustrated in Figure 1a of the main text. Physics-informed PDE solvers can be categorized into domain-specific solvers \cite{raissi2019physics, raissi2017physicsI, raissi2017physicsII} and operator (general function-to-function) solvers \cite{li2023physicsinformed, wang2021learning}. In this section, we benchmark physics-informed domain-specific solvers on full-sized problems.

\begin{figure*}[!ht]
      \centering
      \makebox[\textwidth][c]{\includegraphics[scale=0.5]{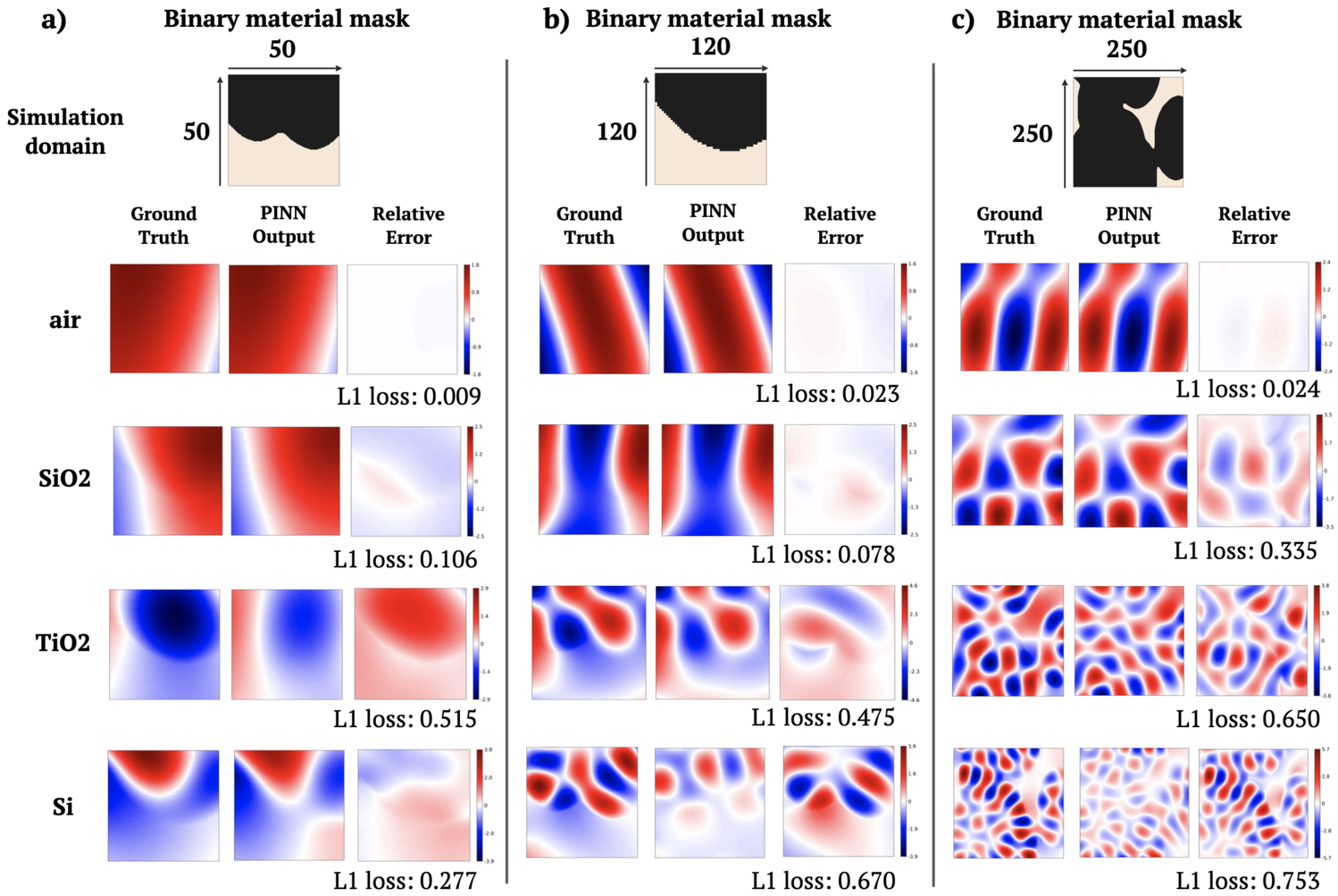}}%
      \caption{PINNs trained on full-sized problems. Here we show the results of training PINNs with sine activation functions on different problem sizes and material types. Three problem sizes are investigated: \textbf{a)} 50 by 50, \textbf{b)} 120 by 120, and \textbf{c)} 250 by 250. For each problem size, a binary mask defines the material geometry, and four different materials with the same geometry are used to construct 4 different problems. We conduct thorough parameter and architecture searches within each problem, and the L1 loss for the best models are reported.
       }\label{SI-S1}
\end{figure*}

\begin{table}[ht!]
\centering
\caption{Range of values for hyperparameter sweep}
\begin{tabular}{ll}
\multicolumn{1}{c|}{\textbf{Hyperparameter}}    & \multicolumn{1}{c}{\textbf{Values}}                           \\ \hline
\multicolumn{1}{l|}{Starting Learning Rate}     & \begin{tabular}[c]{@{}l@{}}min: 1e-5\\ max: 1e-3\end{tabular} \\ \hline
\multicolumn{1}{l|}{ADAM Weight Decay}          & \begin{tabular}[c]{@{}l@{}}min: 1e-8\\ max: 1e-3\end{tabular} \\ \hline
\multicolumn{1}{l|}{Activation Multiplier (w0)} & \begin{tabular}[c]{@{}l@{}}min: 1\\ max: 30\end{tabular}      \\ \hline
\multicolumn{1}{l|}{No. Hidden Layers}          & \begin{tabular}[c]{@{}l@{}}min: 2\\ max: 5\end{tabular}       \\ \hline
\multicolumn{1}{l|}{Layer Height}               & \{32, 64, 128, 256\}                                          \\ \hline
\multicolumn{1}{l|}{B.C. Weight}                & \begin{tabular}[c]{@{}l@{}}min: 1\\ max: 20\end{tabular}      \\ \hline
                                                &                                                              
\end{tabular}
\label{tab:hyperparameters}
\end{table}

\begin{figure}[!ht]
      \centering
      \makebox[\textwidth][c]{\includegraphics[scale=0.7]{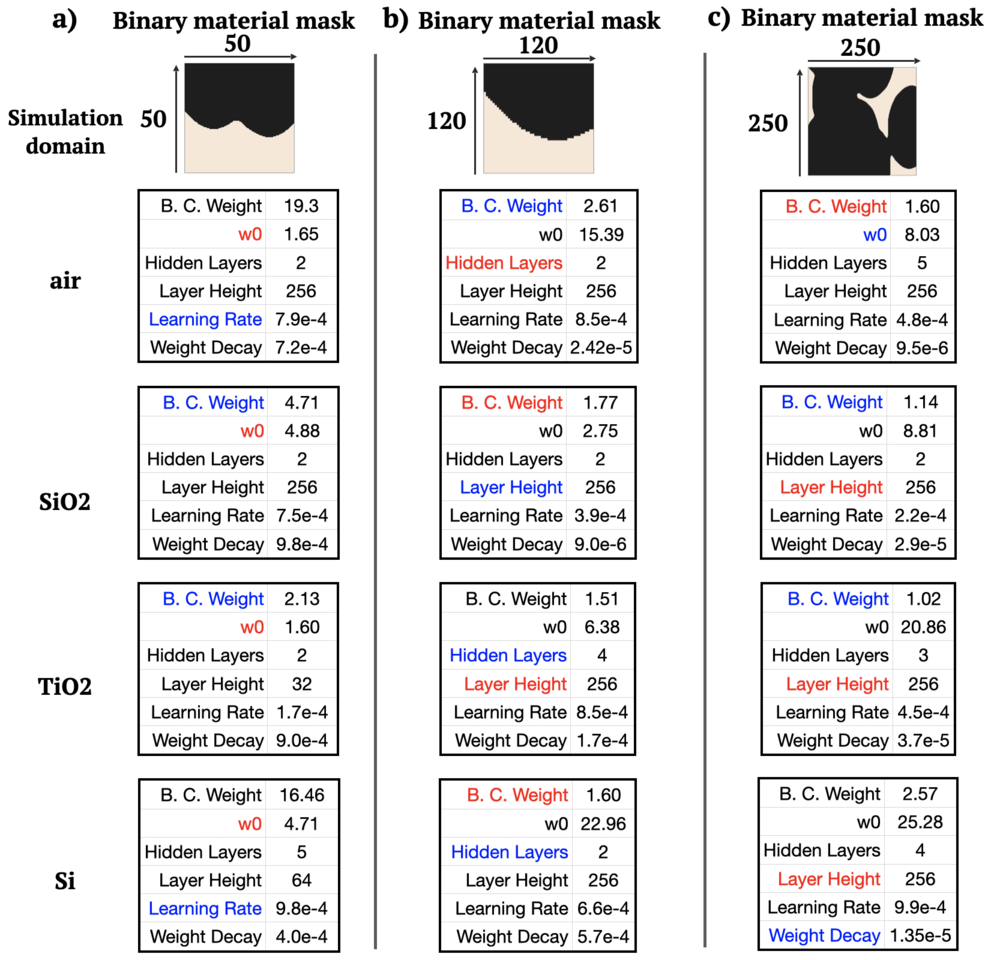}}%
      \caption{Optimal training hyperparameters for each simulation problem, determined as a result of individual Bayesian sweeps. As determined using a multivariate importance score calculation over all hyperparameters, red denotes the most sensitive hyperparameter, and blue the second-most.
       }\label{SI-S1-2}
\end{figure}

We benchmark a fully connected PINN architecture with sine activation \citep{song2022versatile} on a total of 12 full-sized problems: \{air, SiO2, TiO2, Si\} $\times$ \{50 by 50, 120 by 120, 250 by 250\}. The simulation domain is heterogeneous (containing either silicon oxide, titanium oxide, or silicon, and air), is surrounded by Robin-type boundary conditions, and the source is located outside of the domain. For each of the 12 problems demonstrated in Fig. \ref{SI-S1}, we perform Bayesian sweeps over the possible values specified in Table \ref{tab:hyperparameters}. This is because each PINN setup is highly sensitive to the unique Robin boundary conditions of each solution, requiring careful tuning of the training hyperparameters. A total of 50 training runs are performed in each of the 12 sweeps, over a sub-range of hyperparameters that was determined as most promising based on the results of a random-selection sweep of the SiO2 120 by 120 simulation problem consisting of 1200 training runs. 

The optimal parameters determined by each Bayesian sweep are recorded in Fig. \ref{SI-S1-2}. The variety of these results illustrates the challenges related to training domain-specific PINNs on general, heterogeneous simulation domains. Convergence is highly sensitive to the selection of the hyperparameters, but the most sensitive set of hyperparameters is not constant across simulation setups. However, as illustrated in Fig. \ref{SI-S1-2}, the most common hyperparameter of high importance to the outcome of the sweep, as determined by the multivariate "importance" calculation \cite{liu1998evaluation}, is the boundary condition weight (i.e., the weighing factor between the bulk loss and the boundary condition loss during training). Applying PINNs to general simulation domains is thus a laborious training process due to the high sensitivity of model performance on non-consistent sets of training hyperparameters.

A general trend emerges in the simulation sweep, as illustrated in Fig. \ref{SI-S1}: as the material refractive index increases (and with it the complexity of the field profile), and the simulation domain size scales up, the performance of the PINN rapidly deteriorates. Although the PINN performs reasonably well for smaller domain sizes with domains consisting of relatively lower refractive index materials, the fully connected model PINN is incapable of scaling to the larger domain sizes solved by SNAP-DDM with the same levels of accuracy.

The physics training aspect of domain-specific solvers is closely related to the training process in operator learning. There is significant progress in physics-informed operator solvers.
Physics-informed neural operators (PINOs) for example, rely on the FNO framework to learn the function-mapping operator by training on both data and PDE constraints at different resolutions. \cite{li2023physicsinformed} Although it is demonstrated to work well for heterogeneous simulation domains and has several interesting properties, such as training on lower-frequency problems and generalizing to higher-frequency sources, PINO is implemented using the FNO as a backbone, which results in difficulties in scaling to higher dimensions. \cite{li2023physicsinformed} PIDeepONets \cite{wang2021learning} is another example of momentous progress in physics-informed operator learning, which biases the output of DeepONets \cite{lu2021learning} towards physically robust solutions. PIDeepONets augments DeepONets by using automatic differentiation over the input variables, similar to PINNs. \cite{wang2021learning} Although these models demonstrate impressive solution accuracy improvements, generalizability, and data efficiency, the models are computationally expensive to train because the training dataset size is a product of the number of input functions and evaluation coordinates. \cite{wang2021learning} The training computational complexity is further complicated by the significant computational graph size increase due to the automatic differentiation of the input parameters, resulting in significantly longer training time compared to DeepONets. PIDeepONets was not able to converge when trained on the Helmholtz equation for the H field, although this can likely be ameliorated by increasing the number of training collocation points and careful tuning of the training hyperparameters with sufficient compute availability. 

\end{document}